\documentclass[conference]{IEEEtran}
\IEEEoverridecommandlockouts

\usepackage{cite}
\usepackage{amsmath,amssymb,amsfonts}
\usepackage{graphicx}
\usepackage{textcomp}
\usepackage{xcolor}
\usepackage{url}
\usepackage{booktabs}
\usepackage{tabularx}
\usepackage{array}
\usepackage{tikz}
\usetikzlibrary{arrows.meta,positioning,calc}

\graphicspath{{figures/}{./}}

\def\BibTeX{{\rm B\kern-.05em{\sc i\kern-.025em b}\kern-.08em
    T\kern-.1667em\lower.7ex\hbox{E}\kern-.125emX}}

\begin{document}

\title{Which Workloads Belong in Orbit? A Workload-First Framework for Orbital Data Centers Using Semantic Abstraction}

\author{\IEEEauthorblockN{Durgendra Narayan Singh}
\IEEEauthorblockA{\textit{Independent Researcher}\\
Twin Falls, ID, USA \\
durgendra@gmail.com}}

\maketitle

\begin{abstract}
Space-based compute is becoming plausible as launch costs fall and data-intensive AI workloads grow. This paper proposes a workload-centric framework for deciding which tasks belong in orbit versus terrestrial cloud, along with a phased adoption model tied to orbital data center maturity.

We ground the framework with in-orbit semantic-reduction prototypes. An Earth-observation pipeline on Sentinel-2 imagery from Seattle and Bengaluru (formerly Bangalore) achieves $\sim99.7$--$99.99\%$ payload reduction by converting raw imagery to compact semantic artifacts. A multi-pass stereo reconstruction prototype reduces 306~MB to 1.57~MB of derived 3D representations (99.49\% reduction). These results support a workload-first view in which semantic abstraction, not raw compute scale, drives early workload suitability.
\end{abstract}

\begin{IEEEkeywords}
space-based compute, orbital data centers, workload selection, latency, downlink, energy
\end{IEEEkeywords}

\section{Introduction}
Lower launch costs, denser satellite constellations, and growing demand for energy-hungry GPU workloads have revived interest in space-based compute~\cite{ref-megaconstellations,ref-spacex-million-sat-odc}. The key issue is not novelty but systems fit: can orbital compute complement terrestrial clouds in a principled way?

Orbital platforms operate with intermittent contact, strong downlink/uplink asymmetry, limited servicing, radiation-driven faults, and different power/thermal regimes. Those constraints change the workload placement landscape. The central question becomes which workloads are intrinsically compatible with orbital constraints, rather than whether orbital compute is feasible at all.

We treat orbital compute as a constrained resource and propose a workload-suitability framework covering EO preprocessing, RF classification, multi-view 3D reconstruction, navigation services, and LLM workloads, supported by two prototypes that quantify downlink reduction on realistic imagery. The aim is a reproducible workload-selection method and a phased capability model, highlighting near-term value from semantic/geometric abstraction that trades bandwidth for computation.
\subsection{Contributions}
This paper makes the following contributions:

\begin{itemize}

\item \textbf{Constraint Formalization:} A first-principles analysis of orbital compute constraints, including propagation latency, intermittent connectivity, energy availability, bandwidth asymmetry, and fault characteristics.

\item \textbf{Workload-Suitability Framework:} A structured scoring matrix with five evaluation criteria (latency tolerance, bandwidth intensity, fault tolerance, data locality, and compute intensity) to assess workload fit for orbital deployment.

\item \textbf{Empirical Validation via Semantic Abstraction:}
Two quantitative case studies provide experimental grounding for the framework demonstrating how in-orbit abstraction reduces downlink requirements. Earth-observation semantic reduction across Seattle and Bengaluru achieves consistent $\sim99.7$--$99.99\%$ bandwidth reduction, while multi-pass stereo depth-proxy reconstruction compresses 306~MB of raw imagery into 1.57~MB (99.49\% reduction), even under limited (11.45\%) usable disparity coverage.

\item \textbf{Phased Capability Model:} A three-phase roadmap linking orbital infrastructure maturity (GPU-only $\rightarrow$ GPU + low-cost power $\rightarrow$ GPU + low-cost power + LISL) to evolving workload suitability.

\end{itemize}

\subsection{Paper Organization}
Section~II reviews related work. Section~III outlines first-principles constraints. Section~IV introduces the workload suitability matrix and scoring guidelines. Section~V presents the matrix results and a phased roadmap for orbital compute. Section~VI reports the EO semantic-reduction prototype. Section~VII outlines multi-view 3D reconstruction as a representative high-compute Earth observation workload. Section~VIII describes implementation and reproducibility, followed by discussion in Section~IX and conclusions in Section~X.

\section{Related Work}
Prior work relevant to orbital compute spans several established areas: onboard processing for bandwidth-constrained Earth observation~\cite{ref-onboard-eo,ref-onboard-ai}; space networking under intermittent contact and asymmetric links, including DTN and protocol/antenna constraints~\cite{ref-dtn-architecture,ref-dtn-bundle,ref-ccsds-space-packet,ref-space-antenna-handbook,ref-picosat-survey}; optical inter-node links that raise throughput~\cite{ref-deep-space-optical-comms}; and edge/mobile offload research on placement under latency, bandwidth, and energy limits~\cite{ref-edge-survey,ref-cloudlet,ref-mec-survey} (mostly assuming terrestrial connectivity). For space-native workloads, satellite stereo and multi-view reconstruction pipelines (e.g., SGM and DSM generation) demonstrate how dense imagery can be compressed into geometric abstractions~\cite{ref-sgm,ref-dsm-mvs,ref-remote-sensing-models}. Proposals for space-based compute and space solar power argue for potential energy and duty-cycle advantages~\cite{ref-megaconstellations,ref-mankins-ssp,ref-glaser-ssp}, while data-center studies document the electricity footprint that motivates alternative supply or efficiency approaches~\cite{ref-koomey-dc-growth,ref-masanet-dc-recalibration,ref-shehabi-dc-usage}. We position orbital compute as a workload-placement problem and contribute an explicit suitability framework linking these threads to concrete selection criteria.
Recent space-systems research emphasizes onboard data reduction and in-orbit signal analysis~\cite{ref-spacecraft-processing,ref-onboard-signal}, while remote-sensing methods increasingly use semantic abstraction (e.g., segmentation and multi-view stereo) to produce compact representations~\cite{ref-rs-segmentation,ref-stereo-3d}. We integrate these strands within a workload-first suitability framing.

\section{First-Principles Constraints of Space-Based Compute}

\subsection{Latency and Connectivity}
Almost all of today\textquotesingle s internet is powered by terrestrial data centers connected by fiber-optic networks. These links have physical latency of approximately 5~$\mu$s/km; cross-continent communication in the United States often has 20--30~ms round-trip time (RTT), while intercontinental RTT can be 120--180~ms.

The latency for space-based compute depends on orbital altitude, line-of-sight, and ground-station access. If there is no line of sight to a ground station, a satellite must wait to transfer data, increasing end-to-end latency. Inter-satellite links (ISLs) can mitigate this at increased system cost. For example, a dawn--dusk sun-synchronous orbit (SSO) deployment at roughly 600--800~km altitude combined with ISLs yields a range of RTT scenarios (excluding processing time).

\begin{figure}[t]
  \centering
  \includegraphics[width=\linewidth]{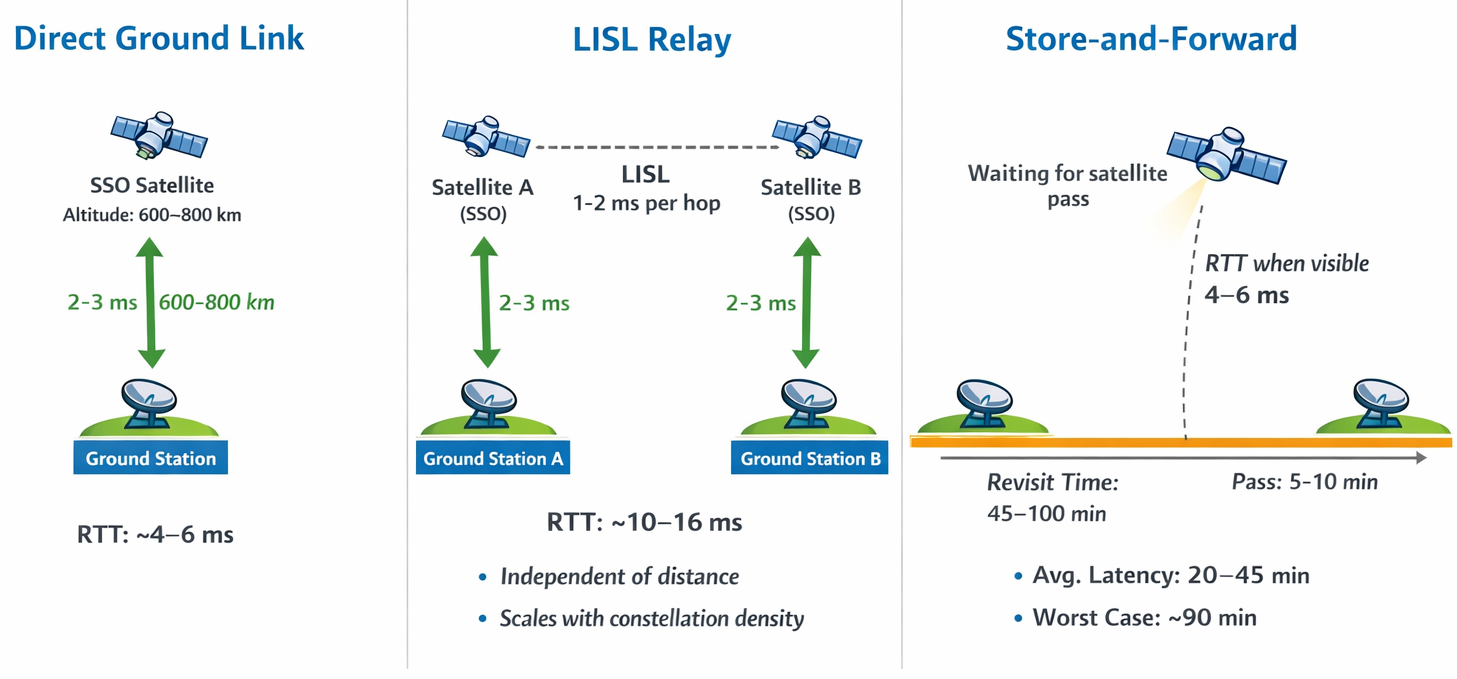}
  \caption{Space--ground connectivity scenarios and latency drivers.}
  \label{fig:latency}
\end{figure}

As illustrated in Fig.~\ref{fig:latency}, while propagation delay for a single hop can be only a few milliseconds, intermittent contact windows and store-and-forward behavior often dominate user-perceived latency for non-continuous coverage architectures.

\subsection{Bandwidth Asymmetry}
Bandwidth asymmetry is driven by downlink limits that lag sensor data growth. In LEO, downlink is constrained by short contact windows, regulated spectrum, and power/antenna limits, while uplink is even more restricted due to command-and-control priority and conservative protocols (often 10:1 downlink-to-uplink or higher). Practically, this makes large uploads and fast feedback loops difficult; operators rely on staged updates and selective downlink. Figure~\ref{fig:bandwidth-symmetry} summarizes this asymmetry at a system level. 


\begin{figure}[t]
  \centering
  \resizebox{\columnwidth}{!}{\begin{tikzpicture}[
  font=\footnotesize,
  box/.style={draw, rounded corners, align=center, inner sep=6pt,
              minimum width=0.47\linewidth, minimum height=3.4cm},
  title/.style={font=\bfseries\footnotesize, text=blue!60!black},
  note/.style={font=\scriptsize, align=center, text width=0.44\linewidth},
  link/.style={-Latex, thick},
  down/.style={text=green!50!black},
  up/.style={text=orange!85!black},
]

\node[box] (L) at (0,0) {};
\node[box, anchor=west, xshift=12mm] (R) at (L.east) {};

\node[title, anchor=north] at ([yshift=-4mm]L.north) {Downlink};
\node[title, anchor=north] at ([yshift=-4mm]R.north) {Uplink};

\node[draw, circle, inner sep=1.6pt, fill=black!10]
  (Lsatn) at ([yshift=10mm]L.center) {};
\node[note, anchor=south] at ([yshift=2mm]Lsatn)
  {SSO satellite (LEO)};

\node[draw, rounded corners, inner sep=2pt, fill=black!5]
  (Lgsn) at ([yshift=-14mm]L.center) {Ground station};

\draw[link] (Lsatn) -- node[midway, right=3pt, down] {$100\,\mathrm{Mbps}$} (Lgsn.north);

\node[draw, circle, inner sep=1.6pt, fill=black!10]
  (Rsatn) at ([yshift=10mm]R.center) {};
\node[note, anchor=south] at ([yshift=2mm]Rsatn)
  {SSO satellite (LEO)};

\node[draw, rounded corners, inner sep=2pt, fill=black!5]
  (Rgsn) at ([yshift=-14mm]R.center) {Ground station};

\draw[link] (Rgsn.north) -- node[midway, right=3pt, up] {$5\,\mathrm{Mbps}$} (Rsatn);

\node[note, text width=0.92\linewidth] at ([yshift=-5mm]$(L.south)!0.5!(R.south)$)
{\textbf{Bandwidth asymmetry:} downlink-to-uplink ratio $\approx 20\times$ (illustrative).};

\end{tikzpicture}}
  \caption{Bandwidth symmetry (illustrative).}
  \label{fig:bandwidth-symmetry}
\end{figure}

\subsection{Energy Availability}
Orbital data centers in dawn--dusk SSO can achieve near-continuous solar exposure (roughly 90--95\% annual capacity factor), enabling functionally continuous power with modest storage. Above the atmosphere, higher solar irradiance and stable incidence angles can yield materially higher energy output than terrestrial solar, while deep space provides a cold radiative sink that reduces cooling overhead. These advantages are best-case and depend on radiation-tolerant design and seasonal shadowing; detailed techno-economic modeling is beyond scope. \cite{ref-megaconstellations}

\subsection{Reliability, Maintenance, and Security}
Orbital platforms are hard to service, so reliability depends on redundancy, graceful degradation, and radiation-tolerant design (e.g., ECC memory and checkpointing). Intermittent uplink limits rapid patching, favoring an immutable-infrastructure posture with signed images, minimal control plane, isolation, and encryption plus audit logs for regulated data.

\section{Workload Suitability Matrix}

\subsection{Purpose of the Matrix}
The matrix provides a structured, repeatable way to evaluate workload fit for space-based compute using measurable criteria.

\subsection{Evaluation criteria and Scoring approach}
Table~II provides details on evaluation criteria, while Tables~III--VII provide scoring logic for each evaluation dimension. Thereafter,

\begin{itemize}
  \item Each workload is scored 1--5 on each dimension.
  \item Higher score means better suitability for space-based compute.
  \item Scores are qualitative but grounded in system behavior.
\end{itemize}

\begin{table}[t]
\caption{Evaluation criteria for workload suitability}
\label{tab:eval-dims}
\centering
\scriptsize
\setlength{\tabcolsep}{4pt}
\begin{tabularx}{\columnwidth}{>{\raggedright\arraybackslash}p{0.22\columnwidth} X}
\toprule
\textbf{Dimension} & \textbf{Meaning (what it captures)} \\ \midrule
Latency tolerance & Sensitivity to added RTT, contact windows, and store-and-forward delays. \\
Bandwidth intensity & Expected reduction from raw inputs to downlinkable outputs; higher reduction improves suitability under downlink constraints. \\
Fault tolerance & Acceptability of retries, partial degradation, and intermittent execution without safety or mission impact. \\
Data locality & Whether the workload’s primary data is generated in space and loses value if moved in raw form. \\
Compute intensity & Computation required per unit of data transferred; higher compute per byte better amortizes constrained links. \\
\bottomrule
\end{tabularx}
\end{table}

\begin{table}[t]
\caption{Latency tolerance scoring (illustrative)}
\label{tab:latency-score}
\centering
\scriptsize
\begin{tabular}{c l l}
\hline
\textbf{Score} & \textbf{Interpretation} & \textbf{Example} \\ \hline
1 & Real-time response ($<1$~s) & Collision avoidance, attitude control \\
2 & Seconds & Dynamic retasking during pass \\
3 & Tens of seconds & Near-real-time alerts \\
4 & Minutes & Batch EO inference \\
5 & Hours+ & Archival analytics \\ \hline
\end{tabular}
\end{table}

\begin{table}[t]
\caption{Bandwidth intensity scoring (illustrative)}
\label{tab:bandwidth-score}
\centering
\scriptsize
\begin{tabular}{c l l}
\hline
\textbf{Score} & \textbf{Interpretation} & \textbf{Example} \\ \hline
1 & Low reduction ($<2\times$) & Raw telemetry relay \\
2 & Minor reduction (2--5$\times$) & Light compression \\
3 & Moderate reduction (5--20$\times$) & Image downsampling \\
4 & High reduction (20--100$\times$) & Segmentation masks \\
5 & Extreme reduction ($>100\times$) & Polygons, object lists \\ \hline
\end{tabular}
\end{table}

\begin{table}[t]
\caption{Fault tolerance scoring (illustrative)}
\label{tab:fault-score}
\centering
\scriptsize
\begin{tabular}{c l l}
\hline
\textbf{Score} & \textbf{Interpretation} & \textbf{Example} \\ \hline
1 & Mission-critical & Onboard control loops \\
2 & Low tolerance & Navigation updates \\
3 & Moderate tolerance & Event detection \\
4 & High tolerance & Image filtering \\
5 & Very high tolerance & Batch ML inference \\ \hline
\end{tabular}
\end{table}

\begin{table}[t]
\caption{Data locality scoring (illustrative)}
\label{tab:locality-score}
\centering
\scriptsize
\begin{tabular}{c l l}
\hline
\textbf{Score} & \textbf{Interpretation} & \textbf{Example} \\ \hline
1 & Earth-originated data & Software updates \\
2 & Mostly terrestrial & Ground sensor fusion \\
3 & Mixed & Cross-domain analytics \\
4 & Primarily space-generated & EO imagery \\
5 & Exclusively space-native & RF spectrum capture \\ \hline
\end{tabular}
\end{table}

\begin{table}[t]
\caption{Compute intensity scoring (illustrative)}
\label{tab:compute-score}
\centering
\scriptsize
\begin{tabular}{c l l}
\hline
\textbf{Score} & \textbf{Interpretation} & \textbf{Example} \\ \hline
1 & Very low & Packet forwarding \\
2 & Low & Format conversion \\
3 & Moderate & Image enhancement \\
4 & High & CNN inference \\
5 & Very high & Multi-stage ML pipelines \\ \hline
\end{tabular}
\end{table}

\subsection{Workload Suitability Matrix}
\begin{equation}
\text{Suitability} \propto 
\frac{\text{Compute Intensity} \times \text{Bandwidth Reduction}}
{\text{Latency Sensitivity}}
\end{equation}
Suitability increases for workloads with high compute intensity and semantic reduction, and decreases with latency sensitivity.
Table~\ref{tab:matrix} shows a mapping of representative workload categories into the five scoring criteria, highlighting which workloads are strong early fits (high bandwidth reduction, high data locality, and delay tolerance) versus those that remain poor fits due to terrestrial data gravity or low reduction. The suitability score is computed as an equal-weighted average across the five evaluation dimensions. Equation (1) provides an intuitive interpretation of this scoring, where suitability increases with compute intensity and bandwidth reduction, and decreases with latency sensitivity. In practice, equal weighting is used to maintain interpretability and avoid overfitting to uncertain system-level parameters. In practice, operators may apply mission-specific weighting depending on whether bandwidth, latency, or power is the primary constraint.
While the scoring criteria are qualitative, they are grounded in first-principles system constraints (propagation delay, contact windows, downlink asymmetry, power duty cycle). The matrix is intended as a structured decision framework rather than a predictive performance model. Scores in Table~\ref{tab:matrix} are assigned based on first-principles system behavior. For example, space RF signal processing scores lower on latency tolerance due to near-real-time detection requirements, but scores high on bandwidth intensity and compute intensity because raw RF data streams are dense and benefit significantly from in-situ classification. Similarly, EO preprocessing scores high on data locality and bandwidth reduction due to space-native data generation and strong semantic compression potential.
 
 \textbf{Summary of recommended workload tiers.} Tier-1 workloads (score $>$ 4.0)—3D reconstruction from satellite imagery, space RF signal processing, and EO preprocessing—are strong early fits because they tolerate additional in-space hops, maximize bandwidth replacement through large semantic reduction, and benefit from pooled orbital compute. These workloads should define product positioning for early space compute companies. Tier-2 / hybrid workloads (score 3.0--3.9)—orbital navigation and timing, small-I/O scientific simulation, batch LLM inference, and LLM training—these workloads are typically more customer-specific and are best positioned as platform extensions rather than anchors.

\section{Phased Capability Roadmap for Orbital Compute}

Orbital compute will mature in phases rather than immediately matching terrestrial hyperscalers. Phase~1 provides onboard GPU acceleration under intermittent downlink, favoring delay-tolerant EO preprocessing and RF classification that convert dense sensor streams into compact artifacts. Phase~2 adds sustained low-cost power, enabling higher-duty-cycle pipelines such as multi-stage inference and multi-view 3D reconstruction, though cross-satellite data movement remains constrained. Phase~3 adds laser inter-satellite links (LISL), enabling cross-node aggregation, multi-pass fusion, global RF mapping, and distributed depth reconstruction before selective downlink.

Table~\ref{tab:phase-workload-legend} summarizes the phase-wise roadmap and workload fit. Practical deployment of orbital compute systems introduces additional constraints beyond workload suitability, including radiation-tolerant hardware, fault-aware workload scheduling, and limited in-situ maintenance. Workloads must be partitioned into restartable units and executed under intermittent connectivity. Integration with existing satellite architectures is expected to follow an edge-processing model, where compute nodes operate alongside payload systems and coordinate across constellations via inter-satellite links.

\begin{table*}[t]
\caption{Workload suitability matrix (scores 1--5 across criteria, illustrative)}
\label{tab:matrix}
\centering
\scriptsize
\setlength{\tabcolsep}{3pt}
\renewcommand{\arraystretch}{1.1}
\begin{tabular}{p{0.23\textwidth}ccccc c}
\toprule
\textbf{Workload category} & \textbf{Lat.} & \textbf{BW} & \textbf{Fault} & \textbf{Local.} & \textbf{Comp.} & \textbf{Avg.} \\ \midrule
3D reconstruction from satellite imagery & 4 & 4 & 4 & 4 & 5 & 4.2 \\
Space RF signal processing & 3 & 5 & 4 & 5 & 5 & 4.4 \\
EO preprocessing (radiometric, geometric) & 5 & 4 & 4 & 5 & 3 & 4.2 \\
Satellite health monitoring / telemetry analytics & 2 & 2 & 2 & 4 & 3 & 2.6 \\
Batch LLM inference & 4 & 2 & 4 & 2 & 4 & 3.2 \\
LLM training & 5 & 1 & 3 & 1 & 5 & 3.0 \\
\bottomrule
\end{tabular}
\end{table*}

\begin{table*}[t]
\caption{Workload suitability and phase-wise roadmap (illustrative)}
\label{tab:phase-workload-legend}
\centering
\scriptsize
\setlength{\tabcolsep}{4pt}
\renewcommand{\arraystretch}{1.08}
\begin{tabularx}{\textwidth}{>{\raggedright\arraybackslash}p{0.28\textwidth} c c c X}
\toprule
\textbf{Workload} & \textbf{P1} & \textbf{P2} & \textbf{P3} & \textbf{Rationale} \\
\midrule
3D reconstruction from satellite imagery &
$\triangle$ & \checkmark & \checkmark\checkmark &
P1: partial steps (tiling, feature extraction, limited depth); P2: cheap power enables full MVS/depth/mesh pipelines; P3: LISL enables in-space aggregation of multi-angle imagery. \\
Space RF signal processing \& classification &
\checkmark & \checkmark\checkmark & \checkmark\checkmark &
P1: dense space-native RF benefits from detection/classification; P2: cheap power enables richer models; P3: LISL enables global spectrum maps and cross-orbit emitter tracking. \\
EO preprocessing (radiometric, geometric) &
\checkmark & \checkmark\checkmark & \checkmark\checkmark &
P1: strong bandwidth reduction from space-native imagery; P2: higher duty cycle enables deeper pipelines and broader coverage; P3: LISL enables cross-satellite fusion and consistent processing. \\
Satellite health monitoring / telemetry analytics &
$\triangle$ & $\triangle$ & $\triangle$ &
P1: useful for autonomy, not bandwidth savings; P2: cheap power does not change limited data-reduction economics; P3: LISL enables fleet-level aggregation but remains secondary. \\
Batch LLM inference &
\texttimes & $\triangle$ & \checkmark &
P1: weak ROI without cheap power and curated inputs; P2: economical for summarization/alerting on down-selected data; P3: aggregates outputs from multiple sources for semantic fusion. \\
LLM training &
\texttimes & \texttimes & $\triangle$ &
P1: power and data gravity make it infeasible; P2: cheap power alone does not solve dataset ingress, hardware refresh, and tooling; P3: may support specialized space-derived training. \\
\bottomrule
\end{tabularx}
\vspace{2pt}

{\scriptsize\raggedright
\textbf{Legend:} $\checkmark\checkmark$ = core/anchor; $\checkmark$ = strong fit; $\triangle$ = opportunistic; $\times$ = not suitable. P1 = GPU only; P2 = GPU + low-cost power; P3 = P2 + LISL.\par}
\end{table*}
\section{EO Semantic Reduction Prototype (Seattle and Bengaluru, 2025)}

\subsection{Method}
We implemented an end-to-end ``raw $\rightarrow$ semantic artifact'' pipeline on Sentinel-2 L2A imagery from two regions (Seattle, USA and Bengaluru, India) in 2025~\cite{ref-pc-stac} to test semantic reduction across distinct cloud regimes.

The workflow uses Sentinel-2\textquotesingle s Scene Classification Layer (SCL) to derive a deterministic cloud mask (cloud/cirrus classes). From the binary mask, it produces two downlink-friendly artifacts: (1) vector cloud polygons from mask contours and (2) ``patch polygons,'' a grid-based tiling that preserves coarse spatial structure with predictable size.

Scenes were grouped by AOI-local cloud fraction into Clear (0--10\%), Mixed (30--60\%), and Cloudy (70--90\%) regimes. Ten scenes per bucket were sampled per region to enable consistent comparisons, mirroring an orbital pattern: filter in orbit, emit small semantic artifacts, and use them to guide routing, prioritization, and selective downlink of raw data.

\subsection{Results: Payload and Downlink Reduction}
Across both regions and cloud regimes, semantic reduction yielded consistently high compression. Aggregate raw volume per ten-scene batch was $\sim 31.46$~MB. Patch polygon payloads were $0.001$--$0.098$~MB by regime, corresponding to $\sim 99.69\%$ to $\sim 99.996\%$ bandwidth reduction.

Cloud morphology remained highly compressible. In cloudy scenes, AOI-local cloud fraction averaged $\sim 80\%$, yet the largest connected deck contained $\sim 95$--$97\%$ of cloudy pixels, so most structure could be represented compactly despite fragmentation. Clear scenes produced near-zero patch counts, while mixed scenes yielded moderate patch complexity, indicating smooth scaling with atmospheric complexity.

Results were consistent between Seattle and Bengaluru despite distinct climatic profiles, with reduction ratios scaling primarily with cloud fraction.
\begin{figure}[t]
  \centering
  \includegraphics[width=\linewidth]{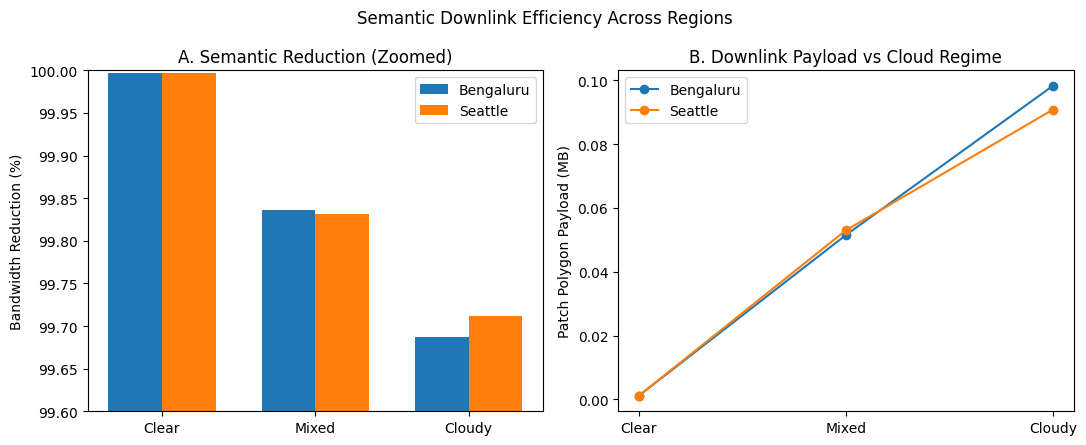}
  \caption{Cross-region semantic abstraction across cloud regimes (Seattle vs Bengaluru, 2025).}
  \label{fig:eo-crossregion}
\end{figure}

\begin{figure}[t]
  \centering
  \includegraphics[width=\linewidth]{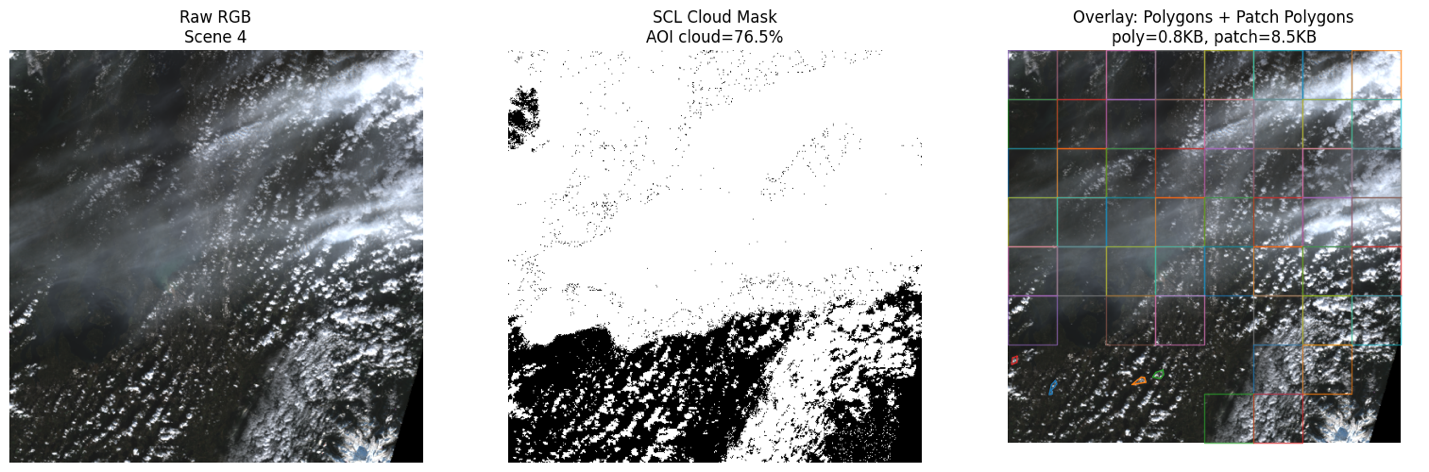}
  \caption{Example EO semantic artifacts (Seattle 2025).}
  \label{fig:eo-semantics}
\end{figure}

\subsection{Orbital Data Center Implication}
At 50~Mbps downlink, a ten-scene batch takes $\sim 5.03$~s raw versus $\sim 0.014$~s as patch polygons. While per-batch savings are modest, they scale linearly with scene volume and constellation throughput. Semantic preprocessing increases effective downlink throughput by roughly $10^2$--$10^4\times$ depending on cloud conditions, therefore, enabling more actionable scenes per pass and reducing dependence on peak windows and ground-station availability.

\subsection{Relation to the Workload Suitability Matrix}
These results validate multiple criteria of the Workload Suitability Matrix. EO preprocessing has high bandwidth intensity (large raw volumes to compact outputs), intrinsic data locality (space-born imagery), and strong latency/fault tolerance (batch processing can be delayed or partial without losing system utility). These characteristics explain why EO preprocessing ranks as a high-suitability workload and reinforce the view that orbital compute derives value by replacing transport with computation when semantic compression is large.

\subsection{Limitations}
This prototype uses deterministic EO preprocessing (Sentinel-2 SCL) on resampled AOI crops and does not model learned inference accuracy, full-resolution tiles, or operational overheads (packetization, encryption, fault tolerance).

\section{3D Reconstruction from Multi-Pass Satellite Imagery (Depth Proxy)} 
\label{sec:depth-proxy}
We evaluate multi-view 3D reconstruction as a representative high-compute workload using publicly available panchromatic ARD (Analysis Ready Data) imagery acquired over the same urban area on different dates. The objective is to assess compute intensity, stereo robustness, and data-reduction characteristics relevant to orbital processing. Imagery is from Vantor (former Maxar Intelligence) Open Data over Los Angeles~\cite{ref-vantor-open-data}; the acquisitions predate the referenced wildfire and are used solely to evaluate reconstruction characteristics.

\subsection{Method (Depth Proxy)}
Depth-proxy reconstruction decomposes into (i) per-pass preprocessing and (ii) cross-pass fusion. Per-pass steps include radiometric normalization, tiling, keypoint/feature extraction, and view-quality scoring. Cross-pass steps include feature matching, multi-view geometry estimation, and depth fusion to produce compact outputs (depth tiles, DSM/DTM, sparse point clouds, or meshes). This staged decomposition matches intermittent downlink and supports progressive refinement: low-resolution depth first, then higher fidelity as additional views arrive.

\subsection{Results: Outputs and Semantic Reduction}
\begin{figure}[t]
  \centering
  \includegraphics[width=\linewidth]{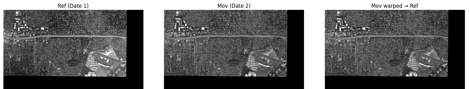}
  \includegraphics[width=\linewidth]{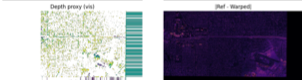}
  \caption{Multi-pass geometric abstraction pipeline (Los Angeles).}
  \label{fig:depth-proxy-phases}
\end{figure}

Feature-based co-registration achieved robust alignment: 6{,}517 matches and 1{,}219 RANSAC inliers (inlier ratio = 0.62) from 20{,}000 ORB keypoints per image. Dense stereo at 3{,}500 $\times$ 3{,}500 (12.25M pixels) produced 11.45\% usable disparity coverage after filtering ($>$1 px), primarily over high-texture urban structures. The two input tiles (305.99~MB) were reduced to 1.57~MB of derived geometric products, a 99.49\% reduction. This shows multi-view reconstruction can transform high-volume raster imagery into compact geometric abstractions suitable for transmission. For context, standard image compression techniques (e.g., JPEG2000 or WebP) typically achieve 5$\times$--20$\times$ reduction on panchromatic imagery, significantly lower than the 99.49\% reduction observed here, indicating that the majority of gains arise from semantic abstraction rather than conventional compression. Figure~\ref{fig:depth-proxy-phases} and Table~\ref{tab:phase-workload-legend} illustrate how phased orbital capabilities enable depth proxies or sparse 3D products to be downlinked in place of full multi-view imagery, reserving raw data for selective verification---a strategy valuable for high-revisit constellations with constrained downlink budgets.

\subsection{Limitations}
This study used multi-date ortho-rectified imagery, so depth estimates are relative proxies. Stereo coverage was limited (11.45\%) by illumination differences and low-texture regions, and the goal is reduction characterization rather than absolute accuracy. Operational 3D reconstruction pipelines typically target coverage above 60--80\% using multi-angle stereo pairs; the observed 11.45\% coverage here reflects the use of multi-date imagery and is intended to demonstrate compute characteristics rather than production-quality reconstruction.

\section{Implementation and Reproducibility}
Both experimental studies were implemented in Python using open satellite datasets and deterministic preprocessing pipelines. Notebooks and scripts are available from the author upon request. All key methodological details and quantitative results are described within the paper.
\section{Discussion}

This paper focuses on workload characterization rather than prescribing orbital architectures. The goal is to formalize workload-placement trade-offs under space-specific constraints and to show, via EO and 3D reconstruction pipelines, that semantic reduction can turn downlink from a fixed bottleneck into a controllable system variable.

Out of scope are detailed economic modeling, radiation-aware benchmarking, hardware reliability analysis, and constellation-scale simulations that include scheduling, networking, and inter-satellite coordination.

\section{Conclusion}

Orbital compute faces unique constraints (latency, intermittent connectivity, downlink asymmetry) but enables in-orbit semantic reduction. This work formalized those constraints and introduced a workload-suitability framework based on latency tolerance, bandwidth intensity, data locality, fault tolerance, and compute intensity.

Empirical prototypes show the value of semantic abstraction: EO semantic reduction achieves $>$99\% downlink reduction across regions, and multi-pass 3D reconstruction reduces 306~MB to 1.57~MB (99.49\%) despite sparse (11.45\%) usable stereo coverage.

These results indicate that delay-tolerant, space-native workloads with high semantic compression are strong early candidates for orbital compute, while tightly interactive or Earth-data--centric workloads remain better suited to terrestrial infrastructure. Workload suitability can be empirically characterized through measurable semantic compression behavior, providing a reproducible basis for future evaluation.

Near-term orbital data center value is therefore not raw compute scale, but transforming space-born data into semantic artifacts before transmission. This workload-first view reframes the optimization objective from moving data to interpreting data.

\section*{Acknowledgment}

This independent work reflects the author’s personal views. The author acknowledges the use of ChatGPT for language refinement and formatting suggestions; all technical content, results, and interpretations remain the author’s responsibility.

\end{document}